# Adaptive Divergence-based Non-negative Latent Factor Analysis

Ye Yuan, Guangxiao Yuan, Renfang Wang, and Xin Luo, *Senior Member, IEEE*

*Abstract*—High-Dimensional and Incomplete (HDI) data are frequently found in various industrial applications with complex interactions among numerous nodes, which are commonly non-negative for representing the inherent non-negativity of node interactions. A Non-negative Latent Factor (NLF) model is able to extract intrinsic features from such data efficiently. However, existing NLF models all adopt a static divergence metric like Euclidean distance or *α-β* divergence to build its learning objective, which greatly restricts its scalability of accurately representing HDI data from different domains. Aiming at addressing this issue, this study presents an Adaptive Divergence-based Non-negative Latent Factor (ADNLF) model with three-fold ideas: a) generalizing the objective function with the *α-β*-divergence to expand its potential of representing various HDI data; b) adopting a non-negative bridging function to connect the optimization variables with output latent factors for fulfilling the non-negativity constraints constantly; and c) making the divergence parameters adaptive through particle swarm optimization, thereby facilitating adaptive divergence in the learning objective to achieve high scalability. Empirical studies are conducted on four HDI datasets from real applications, whose results demonstrate that in comparison with state-of-the-art NLF models, an ADNLF model achieves significantly higher estimation accuracy for missing data of an HDI dataset with high computational efficiency.

## I. INTRODUCTION

A HIGH-DIMENSIONAL AND INCOMPLETE (HDI) matrix [1-4, 7-9, 13-17, 36, 47] is commonly adopted to describe specific relationships among numerous nodes in big-data-related applications like user-item interactions in a recommender system [1-4, 11, 12] and user trustness in a social network services system [5-8]. In general, Owing to the impossibility of maintaining the full connections among involved nodes (e.g., it is impossible to make a user interact with all the others in a large-scale social network services system like Wechat [9, 10]), an HDI matrix commonly has only a few known entries (describing the observed interactions) while the most others are unknown rather than zeroes (describing the unobserved ones).

Although an HDI matrix is highly incomplete, it has rich knowledge regarding desired patterns like users' potential favorites in a recommender system [18, 19]. Hence, a data analytics model is highly needed to acquire such knowledge from an HDI matrix. According to prior research [3, 14, 21-23], a latent factor (LF) model is highly efficient for extracting essential features from an HDI matrix. However, it mostly fails to fulfill the non-negativity constraints, which are desired in describing the inherent non-negativity of node interactions [24-26, 39

A Non-negative Matrix Factorization (NMF) model [27-33, 35] is ubiquitously adopted to represent a full matrix filled with non-negative data. However, they are either incompatible with an HDI matrix [14-17], or yield unnecessarily high cost in computation and storage to achieve such compatibility [34, 35]. In order to perform non-negative latent factor analysis on an HDI matrix conveniently and efficiently, Luo *et al.* [41] propose a Single Latent Factor-dependent, Non-negative and Multiplicative Update (SLF-NMU) algorithm to build a Non-negative Latent Factor (NLF) model. Due to the data density-oriented characteristic of an SLF-NMU algorithm, an NLF model's computational and storage cost is nearly linear with the known entry count of an HDI matrix, which is significantly lower than an NMF model [27-35].

To improve the convergence rate of an NLF model, Luo *et al.* [40] further incorporates the principle of an alternating-direction-method of multipliers into it, thereby achieving an Alternating-direction-method-based Non-negative Latent Factor (ANLF) model. Moreover, to make an NLF model compatible with general and efficient learning algorithms such as stochastic gradient descent (SGD) [45], Luo *et al.* [43] further propose an Inherently Non-negative Latent Factor (INLF) model that adopts a non-negative bridging function to ensure the non-negativity of the output latent factors, thereby making its optimization process compatible with a general algorithm like SGD. the above NLF models' learning schemes must be designed with care for model non-negativity. To date, the above mentioned NLF models are widely adopted in various applications for HDI data representation learning [2, 14, 15].

⬦ Y. Yuan and X. Luo are with the College of Computer and Information Science, Southwest University, Chongqing 400715, China (e-mail: yuanyekl@gmail.com, luoxin21@gmail.com).
⬦ Guangxiao Yuan is with the Business School, University of Technology Sydney, PO Box 123, Broadway NSW 2007, Australia (e-mail: Guangxiao.Yuan-1@student.uts.edu.au).
⬦ R. Wang is with the College of Big Data and Software Engineering, Zhejiang Wanli University, Ningbo 315100, China (e-mail: renfang_wang@126.com).

In spite of their high efficiency in addressing an HDI matrix, existing NLF models [2, 40, 41, 43] all adopt a static divergence metric, like the Euclidean distance [20] or $\alpha$-$\beta$-divergence [2]. As discussed in [2, 14], an NLF model's representation learning ability to an HDI matrix can be improved via tuning the divergence parameters that affects the expression of $\alpha$-$\beta$-divergence. However, the divergence parameters are data-dependent and should be manual-tuned carefully to achieve such positive results. From this point of view, we have encountered the research question:

**RQ.** Is it possible to achieve an adaptive divergence-based NLF model whose divergence metric is self-adaptive?

To answer this question, this paper proposes an <u>A</u>daptive <u>D</u>ivergence-based <u>N</u>on-negative <u>L</u>atent <u>F</u>actor (ADNLF) model with the following three-fold ideas: a) it adopts a generalized objective function with the $\alpha$-$\beta$-divergence, thereby expanding its potential of representing various HDI data, b) it facilitates a non-negative bridging function to connect the optimization variables with output latent factors, thereby achieving a flexible training process compatible with a general learning algorithm as well as keeping model non-negativity constantly, and c) it makes its divergence parameters adaptive through particle swarm optimization, thereby establishing adaptive divergence in its learning objective to achieve high scalability on HDI data from various domains.

This paper mainly contributes in the following perspectives: a) it presents an ADNLF model, which can perform highly-accurate representation learning to HDI data from various domains, and b) Empirical studies on eight HDI matrices testify the excellent performance of the proposed ADNLF model in comparison with state-of-the-art models.

Section II states the preliminaries. Section III presents the methodology. Section IV conducts the empirical studies. Finally, Section V concludes this paper.

## II. PRELIMINARIES

As our fundamental data source, an HDI matrix is defined as [14-17]:

**Definition 1:** Given two large node sets $U$ and $I$, let $R^{|U|\times|I|}$ quantify a certain kind of interactions between them with each entry $r_{u,i}$ denoting the interaction between nodes $u \in U$ and $i \in I$. Let $\Lambda$ and $\Gamma$ denote $R$'s known and unknown node sets, $R$ is an HDI matrix if $|\Lambda| \ll |\Gamma|$. An NLF model tries to build a low-rank approximation to an HDI matrix, which is defined as follows:

**Definition 2.** Given $R$ and $\Lambda$, an NLF model seeks for a rank-$f$ approximation $\tilde{R}$ to $R$ based on $\Lambda$ with $\tilde{R}=PQ^T$ as $P^{|U|\times f}$, $Q^{|I|\times f} \geq 0$.

For such purposes, a Euclidean distance-based objective is:

$$\varepsilon = \sum_{r_{u,i} \in \Lambda} \varepsilon_{u,i} = \sum_{r_{u,i} \in \Lambda} \left( \left(r_{u,i} - \tilde{r}_{u,i}\right)^2 + \lambda\left(\|p_{u,\cdot}\|_2^2 + \|q_{i,\cdot}\|_2^2\right) \right), \quad (1)$$
$$s.t. \ \forall u \in U, i \in I : p_{u,\cdot} \geq 0, q_{i,\cdot} \geq 0,$$

where $p_{u,\cdot}$ and $q_{i,\cdot}$ denote the $u$-th and $i$-th row LF vectors of $P$ and $Q$, $\tilde{r}_{u,i} = p_{u,\cdot} q_{i,\cdot}^T$ denotes the estimate to $r_{u,i} \in \Lambda$ by the NLF model, $\varepsilon_{u,i}$ denotes the instant loss on the training instance $r_{u,i} \in \Lambda$, $\|\cdot\|_2$ computes the $L_2$ norm of an enclosed vector, and $\lambda$ is regularization coefficient, respectively.

## III. METHODS

### A. Generalized Objective Function

As discussed in [44], $\alpha$-$\beta$-divergence is a generalized divergence parameterized by the divergence parameters $\alpha$ and $\beta$. By adjusting the divergence parameters, a wide range of divergences can be achieved. With it, (1) is extended as:

$$\varepsilon = \sum_{r_{u,i} \in \Lambda} \left( -\frac{1}{\alpha\beta}\left(r_{u,i}^\alpha \tilde{r}_{u,i}^\beta - \frac{\alpha}{\alpha+\beta} r_{u,i}^{\alpha+\beta} - \frac{\beta}{\alpha+\beta} \tilde{r}_{u,i}^{\alpha+\beta}\right) + \lambda\left(\|p_{u,\cdot}\|_2^2 + \|q_{i,\cdot}\|_2^2\right)\right), \ s.t. \ \alpha, \beta > 0; \ \forall u \in U, i \in I : p_{u,\cdot} \geq 0, q_{i,\cdot} \geq 0. \quad (2)$$

Note that when $\alpha=\beta=1$ in (2), (1) can be achieved. Hence, the Euclidean distance-based objective (1) is actually a special case of the $\alpha$-$\beta$-divergence-based objective (2). Moreover, the border conditions of $\alpha=0$ and/or $\beta=0$ are not considered in (2) for preventing the divided-by-zero error.

### B. Non-negative Bridging Function

Note that $P$ and $Q$ in (2) play the roles of output LF matrices and optimization parameters simultaneously. Therefore, the learning scheme should be designed with care to fulfill the non-negativity constraints like an SLF-NMU-based one. Nonetheless, such a specifically-designed learning scheme is not compatible with a general and efficient learning algorithm like SGD. This bottleneck can be overcome by incorporating a non-negative bridging function into the objective, thereby separating the constraints from the optimization variables:

$$\forall u \in U, i \in I : p_{u,\cdot} = g(x_{u,\cdot}) \geq 0, \ q_{i,\cdot} = g(y_{i,\cdot}) \geq 0; \quad (3)$$

where $g(\cdot)$ denotes a non-negative bridging function, $x_{u,\cdot}$ and $y_{i,\cdot}$ denote the optimization variables corresponding to $p_{u,\cdot}$ and $q_{i,\cdot}$, respectively. By substituting (3) to (2), we reformulate the objective as:

$$\forall u \in U, i \in I : p_{u,\cdot} = g(x_{u,\cdot}), q_{i,\cdot} = g(y_{i,\cdot}), \tilde{r}_{u,i} = p_{u,\cdot} q_{i,\cdot}^T, \quad (4a)$$

$$\varepsilon = \sum_{r_{u,i} \in \Lambda} \left( -\frac{1}{\alpha\beta}\left( r_{u,i}^\alpha \tilde{r}_{u,i}^\beta - \frac{\alpha}{\alpha+\beta} r_{u,i}^{\alpha+\beta} - \frac{\beta}{\alpha+\beta} \tilde{r}_{u,i}^{\alpha+\beta} \right) + \frac{\lambda}{2}\left( \|p_{u,\cdot}\|_2^2 + \|q_{i,\cdot}\|_2^2 \right) \right), \quad s.t.\ \alpha, \beta > 0; \tag{4b}$$

In (4), the optimization is taken with the optimization variables $X$ and $Y$, and the output LF matrices $P$ and $Q$ are guaranteed to be non-negative through $g(\cdot)$ as in (3). Thus, the optimization process is no long subject to the non-negativity constraints and becomes comparable with a general learning algorithm like SGD [24-26, 46].

Note that the bridging function should be carefully chosen to ensure the performance of a resultant model. According to prior research [37], a sigmoid function can ensure a model's high representation learning ability to non-linear patterns in HDI data. However, a sigmoid function is constantly positive and unable to describe the soft-margins as discussed in [38]. Therefore, we further introduce a boundary threshold into the bridging function for representing such margins as follows:

$$g(z) = \begin{cases} \phi(z) = \dfrac{1}{1+e^{-z}}, \text{if } \phi(z) \geq \iota, \\ 0, \text{if } \phi(z) < \iota; \end{cases} \tag{5}$$

where $\iota$ denotes the threshold constant. With (5), when an output LF becomes small enough during the optimization process, it is then set at zero to describe the soft-margins between involved nodes, thereby boosting the representation learning ability of a resultant model. In general, we set $\iota$ at $5\times 10^{-5}$ empirically.

*C. Learning Scheme*

We apply SGD to (4a) with $X$ and $Y$ for high efficiency:

$$\arg\min_{X,Y}^{SGD} \varepsilon \Rightarrow \forall r_{u,i} \in \Lambda : \begin{cases} x_{u,\cdot}^{(\tau)} \leftarrow x_{u,\cdot}^{(\tau-1)} - \eta \cdot \dfrac{\nabla \varepsilon_{u,i}}{\nabla p_{u,\cdot}^{(\tau-1)}} \cdot \dfrac{\nabla p_{u,\cdot}^{(\tau-1)}}{\nabla x_{u,\cdot}^{(\tau-1)}} \\ y_{i,\cdot}^{(\sigma)} \leftarrow y_{i,\cdot}^{(\sigma-1)} - \eta \cdot \dfrac{\nabla \varepsilon_{u,i}}{\nabla q_{i,\cdot}^{(\sigma-1)}} \cdot \dfrac{\nabla q_{i,\cdot}^{(\sigma-1)}}{\nabla y_{i,\cdot}^{(\sigma-1)}} \end{cases} \Rightarrow \begin{cases} x_{u,\cdot}^{(\tau)} \leftarrow x_{u,\cdot}^{(\tau-1)} + \eta \left( \dfrac{q_{i,\cdot}^{(\sigma-1)}\left(r_{u,i}^\alpha - \tilde{r}_{u,i}^\alpha\right)\tilde{r}_{u,i}^{\beta-1}}{\alpha} - \lambda p_{u,\cdot}^{(\tau-1)} \right) \cdot \dfrac{\nabla p_{u,\cdot}^{(\tau-1)}}{\nabla x_{u,\cdot}^{(\tau-1)}}, \\ y_{i,\cdot}^{(\sigma)} \leftarrow y_{i,\cdot}^{(\sigma-1)} + \eta \left( \dfrac{p_{u,\cdot}^{(\tau-1)}\left(r_{u,i}^\alpha - \tilde{r}_{u,i}^\alpha\right)\tilde{r}_{u,i}^{\beta-1}}{\alpha} - \lambda q_{i,\cdot}^{(\sigma-1)} \right) \cdot \dfrac{\nabla q_{i,\cdot}^{(\sigma-1)}}{\nabla y_{i,\cdot}^{(\sigma-1)}}; \end{cases} \tag{6}$$

where $\eta$ is the learning rate, $\tau$ and $(\tau-1)$ denote the $\tau$-th and $(\tau-1)$-th update points for $x_{u,\cdot}$, $\sigma$ and $(\sigma-1)$ denote $\sigma$-th and $(\sigma-1)$-th update points for $y_{i,\cdot}$, respectively. By combining (5)-(6), we have the following deduction:

$$\begin{aligned} \dfrac{\nabla p_{u,\cdot}^{(\tau-1)}}{\nabla x_{u,\cdot}^{(\tau-1)}} &= \left[ \dfrac{\partial g\left(x_{u,k}^{(\tau-1)}\right)}{\partial x_{u,k}^{(\tau-1)}} \right]_{k=1\sim f} \Rightarrow \dfrac{\partial g\left(x_{u,k}^{(\tau-1)}\right)}{\partial x_{u,k}^{(\tau-1)}} = \begin{cases} \phi\left(x_{u,k}^{(\tau-1)}\right)\left(1-\phi\left(x_{u,k}^{(\tau-1)}\right)\right), \text{if } \phi\left(x_{u,k}^{(\tau-1)}\right) \geq \iota, \\ 0, \text{if } \phi\left(x_{u,k}^{(\tau-1)}\right) < \iota; \end{cases} \\ \dfrac{\nabla q_{i,\cdot}^{(\sigma-1)}}{\nabla y_{i,\cdot}^{(\sigma-1)}} &= \left[ \dfrac{\partial g\left(y_{i,k}^{(\sigma-1)}\right)}{\partial y_{i,k}^{(\sigma-1)}} \right]_{k=1\sim f} \Rightarrow \dfrac{\partial g\left(y_{i,k}^{(\sigma-1)}\right)}{\partial y_{i,k}^{(\sigma-1)}} = \begin{cases} \phi\left(y_{i,k}^{(\sigma-1)}\right)\left(1-\phi\left(y_{i,k}^{(\sigma-1)}\right)\right), \text{if } \phi\left(y_{i,k}^{(\sigma-1)}\right) \geq \iota, \\ 0, \text{if } \phi\left(y_{i,k}^{(\sigma-1)}\right) < \iota. \end{cases} \end{aligned} \tag{7}$$

With (6) and (7), we arrive at the learning scheme of our model.

*D. Adaptive Divergence*

We adopt the principle of <u>P</u>article <u>S</u>warm <u>O</u>ptimization (PSO) [48] to make these four hyper-parameters, i.e., $\alpha_j$, $\beta_j$, $\eta_j$ and $\lambda_j$, adaptive. We start with building a swarm of $q$ particles in a four-dimensional space, where the $j$-th particle maintains an individual group of hyper-parameters. Following the PSO principle, the evolution rule of the $j$-th particle at the $t$-th iteration is formulated as:

$$\forall j \in \{1,\ldots,q\} : \begin{cases} v_j^t = w v_j^{t-1} + c_1 r_1\left(pb_j^{t-1} - s_j^{t-1}\right) + c_2 r_2\left(gb^{t-1} - s_j^{t-1}\right), \\ s_j^t = s_j^{t-1} + v_j^t, \end{cases} \tag{8}$$

where $pb_j^{t-1}$ and $gb^{t-1}$ denote the best position of the $j$-th particle and the whole swarm in the $(t-1)$-th evolving iteration. Note that each particle's position and velocity should be bounded to prevent jumping out of the searching space [48]. $\forall j \in \{1,\ldots,q\}$, $(\alpha_j, \beta_j, \eta_j, \lambda_j)$ is limited in the range of $([\hat{\alpha}, \breve{\alpha}], [\hat{\beta}, \breve{\beta}], [\hat{\eta}, \breve{\eta}], [\hat{\lambda}, \breve{\lambda}])$, $(v_{j\alpha}, v_{j\beta}, v_{j\eta}, v_{j\lambda})$ is limited in the range of $([\hat{v}_\alpha, \breve{v}_\alpha], [\hat{v}_\beta, \breve{v}_\beta], [\hat{v}_\eta, \breve{v}_\eta], [\hat{v}_\lambda, \breve{v}_\lambda])$, respectively. We set $\breve{v} = 0.2 \times (\breve{s} - \hat{s})$ and $\hat{v} = -\breve{v}$ as described in [48].

Note that in (8), each particle determines its next move based on $pb_j$ and $gb$, which are updated as:

$$pb_j^t = \begin{cases} pb_j^{t-1}, F\left(s_j^t\right) \leq F\left(pb_j^{t-1}\right), \\ s_j^t, \quad F\left(s_j^t\right) > F\left(pb_j^{t-1}\right); \end{cases} \quad gb^t = \begin{cases} gb^{t-1}, F\left(s_j^t\right) \leq F\left(gb^{t-1}\right), \\ s_j^t, \quad F\left(s_j^t\right) > F\left(gb^{t-1}\right). \end{cases} \tag{9}$$

where $F(\cdot)$ denotes the fitness function defined as:

$$A(s_0^t)=A(s_q^{t-1}), \ F(s_j^t)=\frac{A(s_j^t)-A(s_j^{t-1})}{A(s_q^t)-A(s_q^{t-1})}; \tag{10}$$

where the quantizing function $A$ quantizes the contribution of each particle to $F$. This study adopts the missing data estimation of an HDI matrix as the representation learning objective. Hence, we formulate the $A$ in the form of root mean squared error (RMSE) as:

$$A = \sqrt{\left(\sum_{r_{u,i} \in \Omega}\left(r_{(j)u,i} - \tilde{r}_{(j)u,i}\right)^2\right)\Big/|\Omega|}, \tag{11}$$

where $|\cdot|_{abs}$ calculates the absolute value of a given value, $\Omega$ represents the validation set, and $\tilde{r}_{(j)u,i}$ is the estimate by the $j$-th particle, respectively. With (8-11), $\alpha$, $\beta$, $\eta$ and $\lambda$ are made to be self-adaptive, thereby achieving adaptive divergence.

## IV. EXPERIMENTAL RESULTS AND ANALYSIS

### A. General Settings

**Evaluation Protocol**. For industrial applications [2, 3, 14, 15], one major motivation to address an HDI matrix is missing data estimation. Note that the RMSE is commonly chosen as the evaluation metric for the estimation accuracy of a tested model:

$$RMSE = \sqrt{\left(\sum_{r_{u,i} \in \Phi} \left|r_{u,i} - \tilde{r}_{u,i}\right|^2\right)\Big/|\Phi|},$$

where $|\cdot|$ calculates the cardinality of an enclosed set, and $\Phi$ denotes testing dataset disjoint with $\Lambda$ and $\Omega$, respectively.

**Datasets.** Four HDI matrices collected by real industrial applications are adopted in our experiments, whose details are summarized in Table I.

TABLE I. Experimental dataset details.

| No. | Name | Row | Column | Known Entries | Density |
|---|---|---|---|---|---|
| D1 | Goodbooks [28] | 53,424 | 10,000 | 5,976,480 | 1.119% |
| D2 | MovieTweetings [11] | 48,953 | 27,911 | 605,126 | 4.4e-2% |
| D3 | Jester [12] | 16,384 | 100 | 1,186,324 | 72.407% |
| D4 | Magazine [42] | 72,098 | 2428 | 89,689 | 5.1e-2% |

Note that known entry set of each HDI matrix is randomly split into 10 disjoint and equally-sized subsets, where seven subsets are chosen as the training set, one as the validation set, and the remaining two as the testing set for 70%-10%-20% train-validation-test settings.

**General Settings**. For achieving the objective results, following general settings are applied to all involved models: a) the LF matrices of each model are initialized with the same randomly generated arrays in each single test on the same dataset to eliminate the initialization bias; b) the dimension of the LF space is set as $f$=20 [14, 15]; c) for ADNLF, the searching space is [0.1, 1.5] for $\alpha$ and $\beta$, and [$2^{-8}$, $2^{-4}$] and [$2^{-7}$, $2^{-3}$] for $\eta$ and $\lambda$, respectively; and d) in PSO, we choose the common parameter settings [48], i.e., $w$=0.729, $q$=10, $c_1$=$c_2$=2, $r_1$ and $r_2 \in (0,1)$ and is generated by a uniform distribution.

### B. Parameter Sensitivity

#### 1) Effects of α and β

In this part of experiments, we call this manually-tuned ADNLF model as the ADNLF-M. Figs. 1 depicts its RMSE as $\alpha$ and $\beta$ vary. Table II records ADNLF-M's lowest RMSE with optimal $\alpha$ and $\beta$ versus its RMSE as $\alpha$=$\beta$=1. From them, we have the following findings:

a) **With appropriate $\alpha$ and $\beta$, ADNLF-M represents an HDI matrix more precisely than the most frequently adopted Euclidean distance.** For instance, as shown in Fig. 1(a) and Table II, on D1, the lowest RMSE is 0.8208 with $\alpha$=1.2 and $\beta$=0.1. However, its lowest RMSE is 0.8262 with $\alpha$=$\beta$=1, i.e., with Euclidean distance. The accuracy gain with the appropriately-tuned $\alpha$=$\beta$-divergence is 0.65%. Similar situations are also encountered on the other datasets.
b) **The optimal $\alpha$ and $\beta$ are data-dependent.** For instance, as shown in Table II, on D3 ADNLF-M achieves the lowest RMSE with $\alpha$=1.2 and $\beta$=0.1. However, on D4, the optimal values are $\alpha$=0.2 and $\beta$=1.5.

TABLE II. ADNLF-M's RMSE with optimal $\alpha$ and $\beta$ V.S. $\alpha$=$\beta$=1.

| Datasets | Best $\alpha$ and $\beta$ | RMSE | RMSE as $\alpha$=$\beta$=1 | Gap |
|---|---|---|---|---|
| D1 | $\alpha$=1.2, $\beta$=0.1 | **0.8208** | 0.8262 | 0.65% |
| D2 | $\alpha$=1.5, $\beta$=0.8 | **1.5171** | 1.5221 | 0.33% |
| D3 | $\alpha$=1.1, $\beta$=1.0 | **1.0062** | 1.0105 | 0.43% |
| D4 | $\alpha$=0.2, $\beta$=1.5 | **1.3272** | 1.3377 | 0.78% |

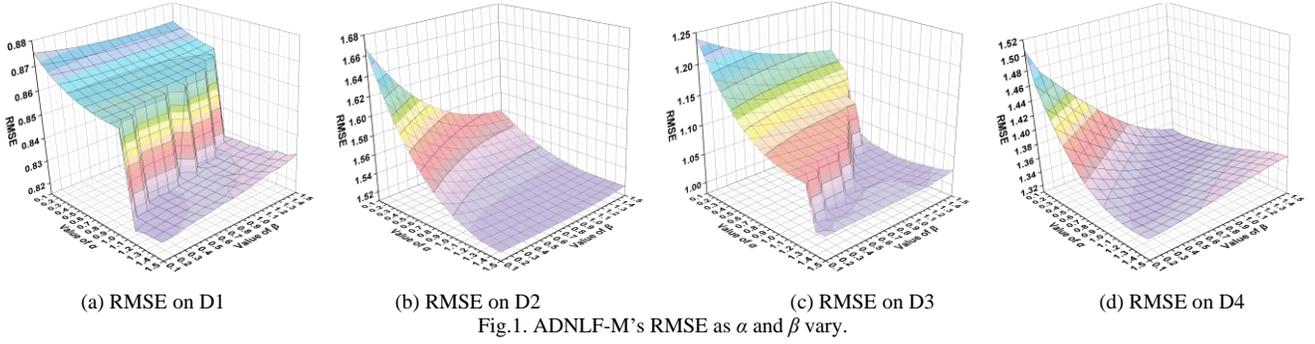

(a) RMSE on D1  (b) RMSE on D2  (c) RMSE on D3  (d) RMSE on D4
Fig.1. ADNLF-M's RMSE as $\alpha$ and $\beta$ vary.

*2) Effects of $\eta$ and $\lambda$*

We fix $\alpha=\beta=1$ to test the effects of $\eta$ and $\lambda$ in this part of experiments. Fig. 2 depicts ADNLF-M's RMSE as they vary. As shown in Figs. 2, the optimal $\eta$ and $\lambda$ in ADNLF-M are also data-dependent, which should be chosen carefully to ensure ADNLF-M's performance. From this part of experiments, we evidently see that the adaptive divergence of ADNLF is vital for its practicability. Without this mechanism, an ADNLF-M model tends to suffer from unstable estimation accuracy for missing data of an HDI matrix. In addition, the divergence and learning parameters should be tuned with care, which results in high computational and time costs. Next, we compare an ADNLF-M model with the optimal hyper-parameters with an ADNLF model with adaptive divergence.

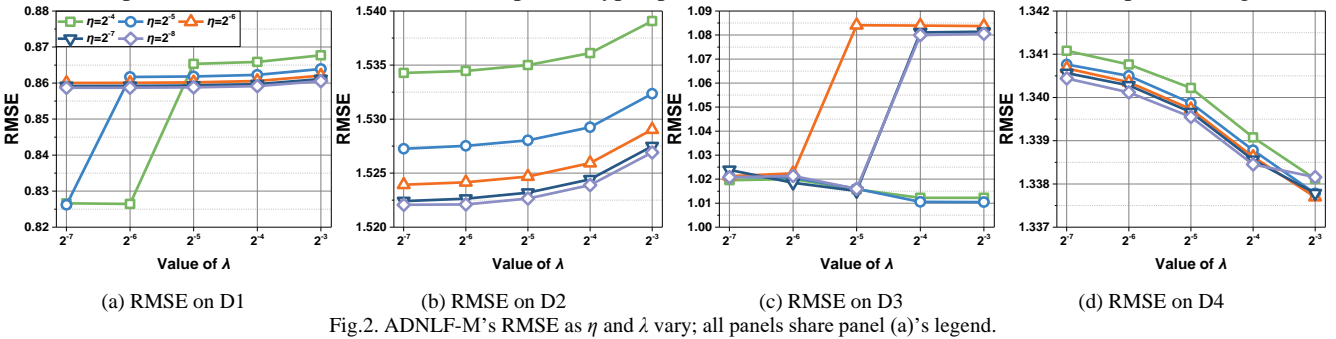

(a) RMSE on D1  (b) RMSE on D2  (c) RMSE on D3  (d) RMSE on D4
Fig.2. ADNLF-M's RMSE as $\eta$ and $\lambda$ vary; all panels share panel (a)'s legend.

## C. Self-adaptation of hyper-parameters

From Figs. 1 and 2, we clearly find that the hyper-parameters, i.e., $\alpha$, $\beta$, $\eta$ and $\lambda$, should be tuned with care to enable the best performance. Therefore, their self-adaptation is highly desired. In this part, we compare the performance of an ADNLF-M and a complete ADNLF models. Tables III compares their performance. These results evidently support the positive effects by the adaptive divergence. More specifically, we have the following findings:

a) **ADNLF's adaptive divergence does not impair its representation learning ability to HDI data.** According to Table III, ADNLF achieves lower RMSE than Manual on D1, D2, D4 and D7. For instance, on D4, ADNLF achieves the lowest RMSE 1.3244 and the lowest RSME of Manual is 1.3272. Thus, self-adaptation of hyper-parameters yields an accuracy gain at 0.32%. This result is quite impressive, since ADNLF's RMSE is achieved with adaptive divergence where the hyper-parameters are all adaptive. In comparison, ADNLF-M's hyper-parameters are all tuned carefully to achieve its lowest RMSE.

TABLE III. Performance comparison between two models.

| Dataset | ADNLF (with Adaptive Divergence) | | | | ADNLF-M (without Adaptive Divergence) | | | |
|---|---|---|---|---|---|---|---|---|
| | **RMSE** | **Iterations** | ***Per** | ****Total** | **RMSE** | **Iterations** | ***Per** | ****Total** |
| D1 | **0.8199** | 11 | 53.16 | **584.76** | 0.8208 | 365 | **5.21** | 1901.65 |
| D2 | **1.5170** | 6 | 1.58 | **9.48** | 1.5171 | 220 | **0.16** | 35.20 |
| D3 | 1.0079 | 23 | 20.52 | **471.96** | **1.0062** | 338 | 2.03 | 686.14 |
| D4 | **1.3244** | 14 | 0.82 | **11.48** | 1.3272 | 355 | **0.08** | 28.40 |

*Time cost per iteration (seconds). **Total time cost (seconds).

Considering D3, D5, D6 and D8, ADNLF's RMSE is slightly higher than that of ADNLF-M. However, according to Table III, the accuracy loss is low. For instance, on D8, ADNLF's RMSE is 1.2232 while ADNLF-M' RMSE is 1.2225, where the accuracy loss is 0.06%.

b) **ADNLF's adaptive divergence greatly improves its computational efficiency.** As shown in Table III, ADNLF converges much faster than ADNLF-M does owing to its adaptive divergence. Thus, compared with ADNLF-M, its total time cost decreases drastically. For instance, as shown in Table III, ADNLF consumes 584.76 seconds to converge on D1, which is about 30.75% of the 1901.65 seconds by ADNLF-M. Same results are achieved on the other datasets.

## D. Comparison against State-of-the-art Models

In this part of the experiments, we compare the proposed ADNLF with several state-of-the-art models on estimation accuracy and computational efficiency. Table IV summarizes the details of all the models. Note that involved LF models, i.e., M1-4, are run on a Tablet with a 2.6 GHz i7 CPU and 16 GB RAM in JAVA SE 7U60. Considering M5-7, due to their computational requirements, they are deployed on a GPU platform with two NVIDIA GeForce RTX 3090 GPU cards. Note that the batch size is set at 128 uniformly for neural network-based models.

Table V records their lowest RMSE and converging total time costs. From them, we have the following important findings:

a) **M1 significantly outperforms its peers in estimation accuracy for missing data of an HDI matrix.** As shown in Table V, M1 can achieve the highest estimation accuracy on D1-8. For instance, on D1, the lowest RMSE of M1 is 0.8199, which is 1.38% lower than 0.8314 by M2, 1.56% lower than 0.8329 by M3, 7.19% lower than 0.8835 by M4, and 0.59% lower than 0.8248 by M5, 12.64% lower than 0.9386 by M6, and 9.39% lower than 0.9049 by M7. Hence, M1's ability to estimate an HDI matrix's missing data is impressive.

b) **M1's computational efficiency is competitive.** Note that the neural network-based models, i.e., M5-7, they perform lots of matrix operations to implement backward propagation (BP)-based training. Therefore, they depend heavily on the GPU-based acceleration. However, as shown in Table V, we clearly see that M1 only consumes more time than M6 on D3. On other datasets, M1's computational efficiency is obviously higher than that of M5-7.

Note that M1's computational efficiency is lower than that of M2-3. The main reason is that M1 frequently calls the function of $f(z)=e^{-z}$ as $z$ be an arbitrary real number as discussed in Sections III-B-C. Compared with M4, M1 achieves higher computational efficiency on all datasets except on D3. However, note that when compared with the time costs M2-4, M1's time cost keeps in the same order of magnitude. In addition, M1 outperforms M2-4 significantly in estimation accuracy for missing data of D1-8. Hence, M1's computational efficiency is competitive.

TABLE IV. Details of compared models.

| Model | Name | Description |
|---|---|---|
| M1 | ADNLF | The proposed model of this study. |
| M2 | NLF | A widely-adopted NLF model based on a SLF-NMU learning scheme [41]. |
| M3 | FNLF | An NLF model that adopts a generalized momentum method [15]. |
| M4 | WNMF | A weighted NMF model [34] based on based on an NMU learning scheme. |
| M5 | I-AutoRec | An auto-encoder paradigm-based LF model [18]. |
| M6 | VAE | An LF model based on the variant of variational autoencoders [22]. |
| M7 | SIF | An LF model based on the simple neural interaction functions [19]. |

TABLE V. Lowest RMSE and their corresponding total time cost (Secs).

| Case | | M1 | M2 | M3 | M4 | M5 | M6 | M7 |
|---|---|---|---|---|---|---|---|---|
| **D1** | RMSE: | **0.8199**±2.3E-3 | 0.8314±2.1E-4 | 0.8329±3.3E-4 | 0.8835±3.5E-5 | 0.8248±2.3E-4 | 0.9386±4.5E-3 | 0.9049±5.2E-3 |
| | Time: | 584.76±7.22 | 134.77±5.56 | **102.87±2.54** | 594.96±5.13 | 6135.25±28.86 | 632±12.83 | 6623.50±214.29 |
| **D2** | RMSE: | **1.5170**±2.3E-4 | 1.8402±5.5E-4 | 1.8562±4.9E-3 | 2.4667±2.6E-3 | 1.9546±3.1E-4 | 1.9259±3.2E-3 | 1.5770±2.7E-3 |
| | Time: | 9.48±1.26 | 5.76±0.04 | **4.09±0.03** | 20.70±1.59 | 763.81±19.31 | 449.43±20.25 | 501.81±13.22 |
| **D3** | RMSE: | **1.0079**±1.4E-4 | 1.0195±3.5E-4 | 1.0176±2.3E-4 | 1.0953±1.8E-4 | 1.0161±9.8E-4 | 1.2419±3.4E-4 | 1.1726±2.2E-4 |
| | Time: | 471.96±27.63 | 27.66±0.52 | **12.90±0.32** | 13.44±0.89 | 510.56±9.04 | 27.87±0.98 | 906.91±3.22 |
| **D4** | RMSE: | **1.3244**±1.1E-3 | 1.4376±1.5E-4 | 1.4286±3.3E-4 | 3.4157±2.9E-4 | 1.4919±1.3E-4 | 1.4839±1.7E-3 | 1.4117±3.7E-4 |
| | Time: | 11.48±1.22 | 4.35±0.23 | **2.21±0.17** | 19.88±2.75 | 680.40±15.08 | 20.63±1.94 | 107.85±4.67 |

## V. CONCLUSIONS

This study proposes an ADNLF model for performing highly accurate representation on an HDI matrix. Compared with the state-of-the-art models, it enjoys a) high representation learning ability to HDI data, and b) high scalability to handle real problems owing to its adaptive divergence.